# Doing Good or Doing Right?
# Exploring the Weakness of Commonsense Causal Reasoning Models

**Mingyue Han[1], Yinglin Wang[2], \***

School of Information Management and Engineering,
Shanghai University of Finance and Economics, Shanghai, China

[1]mingyue.han@163.sufe.edu.cn, [2]wang.yinglin@shufe.edu.cn

## Abstract

Pretrained language models (PLM) achieve surprising performance on the Choice of Plausible Alternatives (COPA) task. However, whether PLMs have truly acquired the ability of causal reasoning remains a question. In this paper, we investigate the problem of semantic similarity bias and reveal the vulnerability of current COPA models by certain attacks. Previous solutions that tackle the superficial cues of unbalanced token distribution still encounter the same problem of semantic bias, even more seriously due to the utilization of more training data. We mitigate this problem by simply adding a regularization loss and experimental results show that this solution not only improves the model's generalization ability, but also assists the models to perform more robustly on a challenging dataset, BCOPA-CE, which has unbiased token distribution and is more difficult for models to distinguish cause and effect.

| A Sample from development dataset |
|---|
| *Premise:* The woman banished the children from her property. |
| *ask-for:* "cause" |
| *Alt1*: The children hit a ball into her yard. × **(effect)** |
| *Alt2:* The children trampled through her garden. √ **(cause)** |

Table 1: A challenging case where BERT predicts wrongly

## 1 Introduction

Supervised learning algorithms recklessly absorbing all the correlations found in training data is statistically correct but might have missed the point (Ahuja et al., 2020). Hence, recent work has focused more on spurious correlations in datasets in computer vision and NLP (Jia and Liang, 2017; McCoy et al., 2019). In inference tasks over natural language, spurious correlation has been identified a lot, such as lexical and grammatical constructs, word overlap, sentence length (Gururangan et al., 2018), and unbalanced token distribution (Poliak et al., 2018; Kavumba et al., 2019). COPA (Roemmele et al., 2011) is a natural language understanding task, which requires a system to choose either a cause or effect of a given story event. It is one of the natural language understanding tasks in SuperGlue benchmark (Wang et al., 2019). Pretrained language models gain a great improvement on COPA, such as BERT (Devlin et al., 2019), RoBERTa (Liu et al., 2019), and ALBERT (Lan et al., 2020). The recent state-of-the-art model on COPA, DeBERTa (He et al., 2020), reached a surprising accuracy of 98.4%. However, the complexity of causal reasoning and the requirements of world knowledge imply that the ability of causal reasoning in PLMs might be overestimated. It is worth exploring whether the models have acquired the ability of causal reasoning.

We observe that 66.8% accuracy can be reached by a text semantic similarity model (Mulyar, 2020) based on BERT which is close to the performance (69.5%) of fine−tuning BERT on COPA training set. It indicates BERT is over-dependent on semantic similarity. Since the *cause* and *effect* of the same event often share the similar context, can PLMs really discriminate what we are asking for? A special case where BERT made mistakes on COPA development set in Table 1 seems to confirm our conjecture. BERT is more likely to fail in these challenging samples where the wrong alternative is the answer of its reverse question type. These investigations imply the models with satisfactory performance might have focused excessively on the topic semantic similarity instead of understanding cause and effect more finely. For this purpose, we design several probing experiments (Section 2) to verify our conjecture: (1) perturbation with distractors, (2) masking question type.

---

\*Corresponding Author

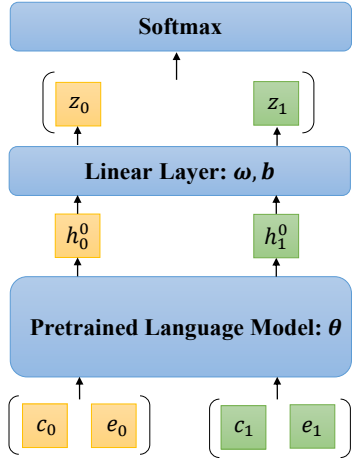

Figure 1: The general architecture of the PLMs on COPA task.

The main work on exploring bias in COPA is from Kavumba et al. (2019). They investigate unbalanced token distributions in correct answers in COPA training set and show that the good performance brought by BERT can be explained by its ability to exploit token distribution in alternatives. They augment the training set with a mirrored-COPA set to prevent the models from predicting with token distribution imprudently. However, we observed this improved model relies on semantic bias more seriously than the original PLMs. We further test the models on a new dataset, BCOPA-CE, which evaluates the ability of a system to distinguish the *cause* and *effect* and to reason without the clues of token distribution. For alleviating the semantic bias problem, we propose to add a regularization loss to the original objective (Section 3). Experimental results show that this solution is not only effective in our challenging test set, but improves the generalization ability of the model on the original test set. It also performs more robustly than the original PLMs in COPA-test *hard* set proposed by Kavumba et al. (2019).

In sum, our contributions are as follows:

(1) We explore the vulnerability of different COPA models by perturbing them with distractive alternatives. (2) We mitigate the weakness of COPA models by adding a regularization loss while maintaining their generalization ability. Our improved models also perform more robustly on the COPA-test hard set. (3) We introduce the BCOPA-CE dataset, which can evaluate the ability of a system to distinguish the cause and effect and to choose cause or effect under unbiased token distribution.

## 2 Probing Experiments

Unlike bias about token distribution or sentence length, indirect semantic cues cannot be analyzed statistically. We explore whether PLMs rely excessively on semantic similarity with special probing experiments. Firstly, we observe whether the model has dropped to a great extent if they see a distractive alternative, like a ***premise***. This distractor cannot be the correct answer, but it has a higher similarity score than the correct alternative. Moreover, inspired by Table 1, we investigate whether the model is aware of the question type during prediction. This is achieved by evaluating the model's performance while removing/masking the question type. We observe whether they still keep good performance without seeing the question type. We describe the model implementation details in Appendix A.

### 2.1 Exp1:Perturbation with Distractors

**Model architecture:** General PLMs assume that the first sentence and the second sentence describe a cause and an effect, respectively. For example, BERT take as input {cause, [SEP], effect}, which entails the question type in its formation. The general architecture in our experiment is shown in Figure 1. The shared parameters $\theta, \omega, b$ are learned to classify each choice independently with the premise, where $(c_i, e_i)$ is the $i$-th cause-effect pair, taking the first hidden vector in the final PLM layer:

$$h_i^0 = \theta(c_i, e_i) \qquad (1)$$

yielding the logits for each cause-effect pair:

$$z_i = \omega^T h_i^0 + b \qquad (2)$$

For training, we pass the logits $[z_0; z_1]$ through a softmax function to determine a probability distribution and minimize the cross-entropy loss with the labels. For prediction, we choose the answer with the highest score by $i^* = argmax_{i \in \{0,1\}} z_i$. If we evaluate the trained models on ternary-choice test set, the prediction is then $i^* = argmax_{i \in \{0,1,2\}} z_i$.

**Perturbation:** We perturb models by adding a third choice, which does not affect human judgment. The "*premise*" is a good candidate since it is highly semantically related to itself while it cannot be the cause or effect of itself due to the non-reflexive trait of causality. We anticipate that the model will change its prediction when it meets the added choice. Meanwhile, we need to make sure that the performance drop is not from the increased difficulty of the problem since it becomes a ternary choice from a binary choice, hence we compare the results with a control experiment, where we add a choice randomly sampled from the COPA-test set.

- **COPA-random**: We control the difficulty of perturbation test by taking a wrong choice randomly sampled from the COPA-test set as the third alternative for each sample. We refer to COPA-random as "**Rand**" in Table 2.
- **COPA-premise:** we take the premise as the third alternative. We refer to COPA-premise as "**Prem**" in Table 2.

### 2.2 Exp2: Masking Question Type

As mentioned above, models are likely to ignore the question information (*cause* or *effect*, often share the same context) if they rely excessively on the semantic similarity. We mask the "*ask-for*" for each sample in COPA-test set by inputting the models with an arbitrary question type. The order of the alternative and the premise is determined by the question type. In masking setting, we randomly input [alternative; premise] or [premise; alternative] for each instance in spite of the question. In this way, half of the samples will keep the original question type, and the other samples get the wrong question type, which do not have the real correct answer. We observe whether these models still keep good performance without seeing the question type. If they do, the question type is ignored for the prediction of the models. We refer to this experimental setting as "**Mask**" in Table 2.

The lower accuracy on "Mask" setting, the more robust the models are.

### 2.3 Baseline models

We conduct the aforementioned experiments with both traditional and SOTA COPA models.

- **CS:** Sasaki et al. (2017) handled the COPA task by statistically estimating causality scores using causal knowledge extracted from a corpus with causal templates.
- **PLMs:** We take BERT-large, RoBERTa-large, ALBERT-xxlarge-v1, and DeBERTa-large as baseline models (referred to as b-l, rb-l, alb, and db-l, respectively), and fine-tune them on the COPA-dev set, using the implementation from hugging face[1].
- **PLMs-aug (**b-l-aug, rb-l-aug, alb-aug and db-l-aug**):** PLMs are fine-tuned on BCOPA, a dataset with unbiased token distribution between the correct alternatives and the wrong alternatives proposed by Kavumba et al. (2019). The BCOPA dataset was constructed by mirroring the original training set with a modified premise.

### 2.4 Results and analysis

As is shown in Table 2, The CS method based on causal knowledge is the most robust system, barely affected by the added alternative. PLMs show different degrees of weakness when they are disturbed by the added alternative. The defensive ability of BERT is the weakest, which is almost completely fooled by distractor and remains the original accuracy without seeing the questions. RoBERTa, ALBERT, and DeBERTa also drop 16.6%, 14.7%, 12.9% respectively compared with the performance of "**Rand**" setting. The fact that the systems perform worse on "**Prem**" (*premise* as a distractor) supports our hypothesis that PLMs have semantic similarity bias. This is because the *premise* is 100% similar to itself, being much more similar than a random distractor. For masking experiments, the theoretical accuracy of a perfectly robust model should be half of the chance-level (i.e., 50%) plus half of the original accuracy. The CS method achieves an accuracy of 61.1% and pays attention to the question type. On the contrary, PLMs seem not to be aware of the question type and perform similarly without this information as original model setting. However, PLMs do not completely ignore the question type since they do not keep the same performance as the original test set.

| Model | Exp1: Perturbation | | | Exp2: Masking | |
|---|---|---|---|---|---|
| | Rand ↑ | Prem ↑ | Δ ↓ | Test ↑ | Mask ↓ |
| CS | 70.1 | 70.5 | -0.4 | 70.8 | 61.1 |
| b-l | 59.3 | 11.6 | 47.6 | 69.5 | 69.0 |
| b-l-aug | 63.3 | 13.0 | 50.3 | 70.0 | 69.6 |
| rb-l | 83.3 | 66.7 | 16.6 | 86.3 | 82.8 |
| rb-l-aug | 85.6 | 65.7 | 19.9 | 87.3 | 83.5 |
| alb | 86.7 | 71.9 | 14.7 | 88.0 | 80.2 |
| alb-aug | 86.4 | 61.2 | 25.2 | 87.9 | 84.1 |
| db-l | 90.8 | 77.9 | 12.9 | 91.6 | 87.8 |
| db-l-aug | 91.1 | 78.9 | 12.2 | 91.8 | 88.8 |

Table 2. The accuracy of models in probing experiments. "↓" denotes a negative indicator (the lower, the better) and "↑" denotes a positive indicator (the higher, the better).

---
[1] The PLMs could be found at
https://github.com/huggingface/transformers

| A Sample in COPA-test set | New Samples in BCOPA-CE test set | |
|---|---|---|
| *Premise:* The accident was my fault.<br>*ask-for:* "effect"<br>*Alt1:* I felt guilty. √<br>*Alt2:* I pressed charges. × | *Premise:* The accident was my fault.<br>*ask-for:* "effect"<br>*Alt1:* I felt guilty. √<br>*Alt2:* I was absent-minded. × | *Premise:* The accident was my fault.<br>*ask-for:* "cause"<br>*Alt1:* I felt guilty. ×<br>*Alt2:* I was absent-minded. √ |

Table 3 The samples in COPA-test set and BCOPA-CE test set.

We also investigate the robustness of the debiased methods of augmenting training data which focus on the unbalanced token distributions proposed by Kavumba et al. (2019). They suffer from the same issue even more seriously than the original PLMs except DeBERTa. This might be due to the fact that the models are more likely to capture the semantic similarity since each alternative pair in BCOPA appears twice.

## 3 Model-improving Method

### 3.1 BCOPA-CE Test

As is shown in Table 3, we introduce a balanced COPA test set, BCOPA-CE, by taking cause event and effect event as two alternatives for each premise. Specifically, for each premise of the 500 samples in COPA-test set, we generate one event manually which is a plausible answer to the opposite question type of the original sample. In the sample in Table 3, for the premise: "*The accident was my fault.*", we generate the *cause* of it: "*I was absent-minded.*", since the original question is asking for "*effect*". After this process, we obtain 500 triplets of <*premise*, *cause*, *effect*>. Then, we construct 1000 samples by giving two different questions (*cause* or *effect*) to each triplet. This guarantees the balanced token distribution between the correct and the wrong alternatives. The dataset generation details are described in Appendix B. Human evaluation has been conducted to ensure the quality of the new dataset in Appendix C.

### 3.2 Regularization Loss

We expect the model to make good choices while paying attention to the question type. For a sample in the COPA training set, the proposed loss includes two parts: The CrossEntropy loss and a regularization loss. The first part prompts the model to answer correctly given the question type. The extra regularization loss requires that a model should be neutral when it sees the opposite question type for the same premise and same alternatives, since neither alternative is the correct answer.

General PLMs take the first input sentence as the cause, and the second sentence as the effect. Mathematically, the logits of two input sentences in reverse cause-effect order should be as close as possible, even if one of two alternatives is semantically similar to the premise (the correct answer of the original question).

$$L = (1 - \lambda) * L_{CE} + \lambda * L_{Reg} \quad (3)$$

$$L_{Reg} = \|z_0^r - z_1^r\|_2^2 \quad (4)$$

$z_i^r$ is the logit of input $[e_i; c_i]$ computed by equation (1), which reverses the order of cause and effect of choice $i$. We set $\lambda = 0.01$ in all experiments corresponding to regularization loss.

### 3.3 Result and Analysis

Table 4 demonstrates the performance of our improved models on the COPA-test set, the BCOPA-CE set and the COPA-hard set. It's noted that the models with a regularization loss not only have improved performance on BCOPA-CE set, but also perform better than the original PLMs on COPA-test set. Previous debiased models on token distribution perform worse than the original model, which is consistent with our conjecture that they amplify the semantic bias. Our solution also performs better on COPA-test-hard than the original PLMs, which has balanced token distribution as Kavumba et al. (2019) introduced. Regularization in our method considers debiasing token distribution as well, because we tend to stop the models from capturing any cues when it reverses the input order.

| Model | Test ↑ | BCOPA-CE ↑ | Δ ↓ | Test-hard ↑ |
|---|---|---|---|---|
| b-l | 69.5 | 51.5 | 18.0 | 61.6 |
| b-l-reg | **71.1** | **64.1** | **7.0** | 63.6 |
| b-l-aug | 70.0 | 51.1 | 18.9 | **69.7** |
| rb-l | 86.3 | 73.0 | 13.3 | 83.1 |
| rb-l-reg | **87.7** | **83.9** | **3.8** | 84.5 |
| rb-l-aug | 87.3 | 69.2 | 18.2 | **87.0** |
| alb | 88.0 | 80.5 | 7.6 | 86.9 |
| alb-reg | **89.4** | **86.7** | **2.7** | **88.6** |
| alb-aug | 87.9 | 71.4 | 16.5 | 88.0 |
| db-l | 91.6 | 72.3 | 19.3 | 88.6 |
| db-l-reg | **92.2** | **86.3** | **5.9** | 89.7 |
| db-l-aug | 91.8 | 69.8 | 21.9 | **90.5** |

Table 4 The performance of PLMs and their variants on challenging set. Bold represents the best model setting in the same PLM.

### 3.4 Error Analysis

We conduct an error analysis for the SOTA model, DeBERTa, using the run that is closest to the average of 20 runs. We give an example (the second row) from the BCOPA-CE dataset in Table 5 where DeBERTa predicts wrongly but the regularized DeBERTa model succeeds. Interestingly, both models make a correct prediction on the original sample (the first row) from COPA-test set, which indicates that the new alternative we generate perturbs the choice of the original DeBERTA model.

We calculate the word importance of all tokens in correct answer through erasure (Li et al., 2017). The importance score is computed by the relative difference in log likelihood on gold-standard labels while replacing the token with [MASK]. We observe two models predict correctly in this original sample but with different attention on tokens. As is shown in Figure 2, DeBERTa chooses Alt2 by focusing on "*He*" and "*spoke*", but DeBERTa-reg pays the most attention to

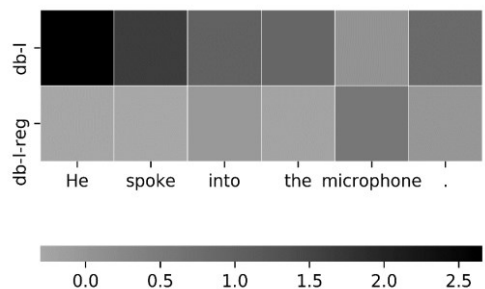

Figure 2: Heatmap of importance of each token in correct answer for the db-l model and db-l-reg model.

| Original sample | *Premise:* The man's voice projected clearly throughout the auditorium. *Ask-for:* cause <br> *Alt1:* He greeted the audience. × <br> *Alt2:* He spoke into the microphone. √ |
|---|---|
| New sample | *Premise:* The man's voice projected clearly throughout the auditorium. *Ask-for:* cause <br> *Alt1:* Everyone heard him. × <br> *Alt2:* He spoke into the microphone. √ |

Table 5 The case where DeBERTa is perturbed but regularized DeBERTa not.

"*microphone*", which is more in line with human causal intuition. When people make such inference, the causal relation between "*microphone*" and "*projected clearly throughout the auditorium*" should be more important than the co-reference relationship.

## 4 Conclusion

In this paper, we explore whether COPA models rely excessively on semantic similarity for prediction. We add the regularization loss to the training objective to alleviate this weakness. Results show that our solution is effective in our adversarial test, and improve the generalization ability and the robustness of models on previous COPA-hard dataset. Moreover, previous debiased models on token distribution rely on semantic bias more seriously than the original models, which reminds us if debiasing bring more other bias.

## Acknowledgements

Special thanks to all annotators for their hard work. This work was supported by the National Natural Science Foundation of China (under Project No. 61375053). We thank Ming Wang, Jingshu Zhang, Dandan Li and the anonymous reviewers for their insightful comments and discussion.

# Appendix

## A Implementation Details

**PLMs:** We randomly split the training set (COPA-dev, or BCOPA) into training set and development set with a ratio of 9:1, and finetune our model up to 20 epochs by implementing an early-stopping strategy with a patience of 5 epochs and using AdamW optimizer. We run 20 different random seeds for each supervised model and report the mean of the non-degenerate runs for each model, which have higher than 80% of accuracy in the training set as in previous work (Niven and Kao, 2019).

**CS:** We reproduce the preprocessing of their work and achieve 70.8% accuracy, which is slightly lower than the reported accuracy of 71.4%.

All parameters are learned from the development set by manual tuning. The best-performing parameter is determined by the accuracy of the model in the development set. The final parameters in our experiments are shown in Table 6.

| Model | LR | BS | WD | WP | $\lambda$ |
|---|---|---|---|---|---|
| b-l | 1e-4 | 32 | 0.01 | 0.1 | - |
| b-l-aug | | | | | - |
| b-l-reg | 8e-5 | | | | 0.01 |
| rb-l | 8e-6 | 32 | 0.01 | 0.06 | - |
| rb-l-aug | | | | | - |
| rb-l-reg | 1.2e-5 | | | | 0.01 |
| alb | 1.1e-4 | 48 | 0 | 0 | - |
| alb-aug | | | | | - |
| alb-reg | 6e-5 | | | | 0.01 |
| db-l | 5e-6 | 32 | 0.01 | 0.06 | - |
| db-l-aug | | | | | - |
| db-l-reg | 1e-5 | | | | 0.01 |

Table 6. The best Batch Size (BS), Learning Rate (LR), Warm up rate (WP), and Weight Decay value (WD) we used in our experiments.

## B Construction Details of BCOPA-CE

We asked five fluent English speakers who have background knowledge of NLP to create the new alternative with the specific guidelines. We instructed creators with requirements of sentence length, overlap rules, and expressions similar to Kavumba et al. (2019).

## C Human Evaluation on BCOPA-CE

We have 1000 samples in BCOPA-CE set, which consist of. 500 samples whose answers are same with original COPA-test set (the left sample in the second column in Table 3, referred to as COPA-CE-ori) and 500 samples whose answers are the choices that we generate (the right sample in the second column in Table 3, referred to as COPA-CE-opp). To ensure the quality of generated dataset, we conduct a quality evaluation with two questions:

- **Q1**: *Are the instances in BCOPA-CE dataset comparable in difficulty to the COPA-test instances?*
- **Q2**: *Is the new alternative we collect plausible for the opposite question type?*

| | COPA-test | COPA-CE-ori | COPA-CE-opp |
|---|---|---|---|
| **Accuracy** | 0.980 | 0.990 | 1.000 |
| **Fleiss' Kappa** | 0.919 | 0.893 | 0.890 |

Table 7: Human evaluation result of generated dataset.

We evaluate the accuracy of human on both COPA-CE-ori dataset and COPA-CE-opp dataset to answer the Q1 and evaluate the human performance on COPA-CE-opp set for Q2. The COPA-CE-opp set changes the question type and takes the generated event as gold answers, hence it can be evaluated for the plausibility of generated alternatives. We asked 9 people to make choices, each group of 3 people for one dataset. We determine the final choice by majority voting. The inter-annotator agreement is calculated by Fleiss' Kappa. As is shown in Table 7, the BCOPA-CE set has comparable difficulty with COPA-test. The performance on COPA-CE-opp shows that the new alternatives we create are plausible.